\documentclass{SCIS}

\usepackage{multirow}
\usepackage{rotating}
\def\method{GPS}

\begin{document}

\ArticleType{RESEARCH PAPER}
\Year{2024}
\Month{}
\Vol{}
\No{}
\DOI{}
\ArtNo{}
\ReceiveDate{}
\ReviseDate{}
\AcceptDate{}
\OnlineDate{}

\newcommand{\revA}[1]{{\color{red} #1}}

\title{GPS: Graph Contrastive Learning via Multi-scale Augmented Views from Adversarial Pooling}{GPS: Graph Contrastive Learning via Multi-scale Augmented Views from Adversarial Pooling}

\author[1]{Wei Ju}{}
\author[1]{Yiyang Gu}{}
\author[1]{Zhengyang Mao}{}
\author[2]{Ziyue Qiao}{}
\author[1]{Yifang Qin}{}
\author[3]{\\Xiao Luo}{xiaoluo@cs.ucla.edu}
\author[2]{Hui Xiong}{}
\author[1]{Ming Zhang}{{mzhang\_cs@pku.edu.cn}}

\AuthorMark{Wei Ju}

\AuthorCitation{Ju W, Gu Y Y, Mao Z Y, et al}


\address[1]{School of Computer Science, National Key Laboratory for Multimedia Information Processing, \\Peking University, Beijing {\rm 100871}, China}
\address[2]{Artificial Intelligence Thrust, The Hong Kong University of Science and Technology, Guangzhou {\rm 511453}, China}
\address[3]{Department of Computer Science, University of California, Los Angeles {\rm 90095}, USA}

\abstract{Self-supervised graph representation learning has recently shown considerable promise in a range of fields, including bioinformatics and social networks. A large number of graph contrastive learning approaches have shown promising performance for representation learning on graphs, which train models by maximizing agreement between original graphs and their augmented views (i.e., positive views). Unfortunately, these methods usually involve pre-defined augmentation strategies based on the knowledge of human experts. Moreover, these strategies may fail to generate challenging positive views to provide sufficient supervision signals. In this paper, we present a novel approach named \textbf{G}raph \textbf{P}ooling Contra\textbf{S}t (\method{}) to address these issues. Motivated by the fact that graph pooling can adaptively coarsen the graph with the removal of redundancy, we rethink graph pooling and leverage it to automatically generate multi-scale positive views with varying emphasis on providing challenging positives and preserving semantics, i.e., strongly-augmented view and weakly-augmented view. Then, we incorporate both views into a joint contrastive learning framework with similarity learning and consistency learning, where our pooling module is adversarially trained with respect to the encoder for adversarial robustness. Experiments on twelve datasets on both graph classification and transfer learning tasks verify the superiority of the proposed method over its counterparts.}

\keywords{Graph Representation Learning, Graph Neural Networks, Graph Contrastive Learning, Graph Augmentations, Graph Pooling}

\maketitle

\section{Introduction}

With the prevalence of graph-structured data~\cite{cao2022geometric,wu2022multi,chen2022cross,wang2018beyond,wang2022graph}, it is vital to develop effective representations of whole graphs for various real-world applications such as protein/molecular property prediction~\cite{jiang2017aptrank,ju2023few}, drug discovery~\cite{kojima2020kgcn,hao2020asgn,rozemberczki2022chemicalx}, traffic forecasting~\cite{qu2020data,fang2021spatial,zhao2023dynamic}, and recommender systems~\cite{qin2023diffusion,qin2023learning,wang2022collaboration}. Graph neural networks have recently emerged as powerful tools for learning graph representations in fully-supervised or semi-supervised scenarios~\cite{ying2018hierarchical,ju2022kgnn,jin2022deepwalk,luo2023towards}. However, obtaining a large number of label annotations is often challenging, particularly in highly specialized domains such as biochemistry~\cite{hao2020asgn}. While the number of labeled graphs may be restricted, unlabeled graphs are quite straightforward to acquire in practice. Hence, plenty of efforts have been directed towards self-supervised graph representation learning, which explores unlabeled graphs to alleviate the dependency on massive label annotations.

Motivated by the recent progress in computer vision~\cite{he2020momentum,chen2020simple} and recommender systems~\cite{zhou2021contrastive,wei2021contrastive,xiao2022lecf}, recent researches attempt to integrate contrastive learning to representation learning in the graph machine learning~\cite{sun2020infograph,chu2021cuco,you2020graph,hassani2020contrastive,wang2021self}. The primary principle underlying graph contrastive learning (GCL) methods is to maximize the Mutual Information (MI)~\cite{linsker1988self} between the input graph and its representation. Specifically, these approaches anticipate that a graph has a representation which is similar to its own augmented view and distinct from other graphs. Thus, these methods can provide discriminative graph-level representations, which are beneficial for a variety of downstream applications.

Despite their superior performance, existing self-supervised methods rely on handcrafted augmentation strategies to provide positive views for comparison. Common strategies include node dropping, edge perturbation, attribute masking, graph diffusion~\cite{hassani2020contrastive} and subgraph~\cite{you2020graph}. These handcrafted strategies, however, have the following drawbacks. 
First, current methods are inconvenient to apply to different datasets since they require expert knowledge to select appropriate strategies for preserving semantics. Edge perturbation, for example, has been empirically demonstrated to benefit social networks but harm certain biological molecules, while node dropping and subgraph are typically beneficial across datasets~\cite{you2020graph}. Moreover, when dealing with datasets from unknown domains, we may require extensive trials to determine the appropriate augmentation strategies, making it inefficient for practical applications. 
Second, these pre-defined strategies could fall short of generating challenging positive views to provide sufficient supervision signals. In particular, we expect that augmented samples can fully discard redundant information from different perspectives, implying the representations of challenging positives are far from those of the original graphs. If the augmented views are close to the original samples, the representation collapse may even occur, resulting in trivial outputs.

Graph pooling is another central area of research for graph representation learning, which is originated from the traditional convolutional neural network for extracting information efficiently. Graph pooling can be divided into topk-based methods~\cite{lee2019self,gao2019graph} and cluster-based methods~\cite{ying2018hierarchical,ranjan2020asap}, which can effectively learn to reduce the redundant information while preserving semantics. Specifically, they either select important nodes from the original graph or group nodes into clusters and coarsen the graph. To sum up, graph pooling has the potential to improve graph contrastive learning since it can adaptively remove the redundancy of the graph from different perspectives.
However, existing researches typically study different pooling manners in supervised scenarios~\cite{sun2021sugar}. 
It is still unclear how to integrate graph pooling methods into graph contrastive learning by automatically providing effective augmented views.

In this paper, we propose a novel approach named \method{} by leveraging learnable graph pooling to generate positive views for effective contrastive learning. Apart from introducing a graph encoder for producing effective graph representations, we involve two graph pooling modules to generate positive views with different emphases on providing challenging positives and preserving semantics, i.e., strongly-augmented view and weakly-augmented view. On the one hand, we directly maximize the similarity of a graph and its weakly-augmented view in a hard manner. On the other hand, we explore the semantics involved in strongly-augmented views by consistency learning between the similarity of two views in a soft manner. Further, our two pooling modules are adversarially trained with the graph encoder for adversarial robustness and efficiency.
Finally, we conduct extensive experiments to empirically validate the effectiveness of our proposed approach \method{}, validating the superiority over state-of-the-art baselines on graph classification and transfer learning tasks.


\section{Related Work}
\label{sec::related}


\noindent\textbf{Graph Representation Learning}
aims to learn effective representations of graph topology and node attributes, which can be categorized into matrix factorization-based, random walk-based, and neural network-based. Matrix factorization-based methods~\cite{ahmed2013distributed,cai2010graph} directly adopt classic techniques for dimension reduction. Random walk-based methods such as DeepWalk~\cite{perozzi2014deepwalk} and node2vec~\cite{grover2016node2vec} model probabilities of co-occurrence pairs using noise-contrastive estimation~\cite{gutmann2010noise}. Neural network-based methods, especially graph neural networks (GNNs), have attracted increasing interest in recent years. 
With the development of representation learning, various GNNs~\cite{fan2020graph,guo2021syntax,ju2023tgnn,mao2023rahnet,wu2022multi,chen2022cross,wang2018beyond,zou2023simple} have achieved state-of-the-art performance. Generally, a GNN shares a common spirit: extracting local structural features by message passing~\cite{gilmer2017neural,ju2023comprehensive} where nodes iteratively aggregate messages from neighboring nodes through edges. With \method{}, besides learning effective graph representations derived from GNNs, we also benefit from graph pooling to automatically generate multi-scale view augmentations.

\smallskip\noindent\textbf{Contrastive Learning on Graphs}
has become a dominant component in self-supervised learning on graphs. Inspired by previous success in visual representation learning, some recent works~\cite{velickovic2019deep,sun2020infograph,zhu2021graph,you2020graph,ju2023glcc,yi2023redundancy} marry the power of contrastive learning and GNNs, and have shown competitive performance. The key idea of these methods is to maximize the agreement between semantics-invariant transformations of the graphs. 
GCA~\cite{zhu2021graph} generates different views by incorporating various priors for graph topology and semantics. GraphCL~\cite{you2020graph} explores the augmentations from the aspects of node dropping, edge perturbation, attribute masking, and subgraph sampling. However, existing works typically involve inflexible and pre-defined augmentation strategies based on the knowledge of human experts, while our approach leverages learnable multi-scale graph pooling to generate positive views for contrastive learning.

\smallskip\noindent\textbf{Graph Pooling}
is a central component of a range of graph neural network architectures~\cite{lee2019self,gao2019graph,ying2018hierarchical,ranjan2020asap,wang2020haar}. It is originated from the traditional convolutional neural networks (CNNs) and reduces the number of parameters of CNNs by downsampling and summarizing from the representations, which makes the training process highly efficient. Similarly, some studies try to generalize pooling operations to graphs for extracting effective information of the whole graph hierarchically, and these graph pooling methods can be boiled down to two categories: TopK-based pooling and cluster-based pooling.

\textbf{TopK-based Pooling} aims to select the most important nodes from the original graph and use these nodes to construct a new graph. SAGPool~\cite{lee2019self} leverages the self-attention mechanism~\cite{vaswani2017attention} to select nodes by considering both node features and graph topology. In gPool~\cite{gao2019graph}, the nodes are selected via mapping the node feature into the importance scores. They share a similar idea to learn a sorting vector based on node representations using GNNs, which indicates the importance of different nodes. 

\textbf{Cluster-based Pooling} tries to utilize an assignment matrix to achieve pooling by assigning nodes to different clusters and coarsen the graph hierarchically. DiffPool~\cite{ying2018hierarchical} treats graph pooling as a node clustering problem and introduces a differentiable pooling module to decide the pooled graph topology. ASAP~\cite{ranjan2020asap} learns a sparse soft cluster assignment for nodes to cluster local subgraphs hierarchically for effectively capturing the graph substructure.

Our framework rethinks the powerful capability of graph pooling and makes the first attempt to leverage learnable graph pooling to derive augmented views in an adversarial manner.


\section{Methodology}

\begin{figure*}
    \centering
    \includegraphics[width=1\textwidth]{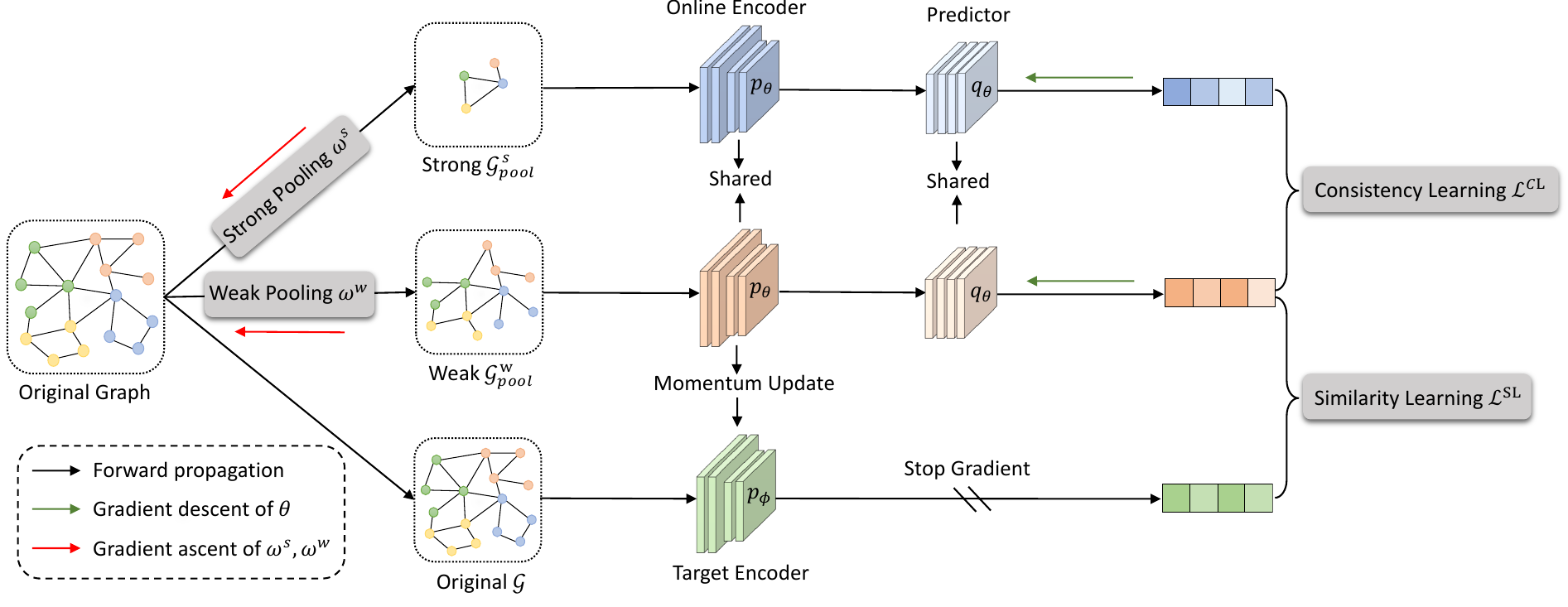}
    \caption{{Illustration of the proposed framework \method{}. We first generate two positive views via our two pooling modules. Then, the two augmented views are fed into the online network while the original graph is fed into the target network. Our contrastive learning framework captures similarity learning and consistency learning, where the graph pooling modules are adversarially trained with respect to the encoder.}}
    \label{fig:framework}
\end{figure*}

In this section, we propose \method{}, a novel graph contrastive learning method, and the overall architecture is shown in Figure \ref{fig:framework}. The positive views play a critical role in graph contrastive learning and deserve a careful design. Previous methods usually generate positive views by handcrafted augmentation strategies, which require expert knowledge and fail to generate challenging positives for providing sufficient supervision signals. To address these problems, we leverage graph pooling techniques to construct positive views with a varying focus on challenging positives and semantic preservation, i.e., strongly-augmented view and weakly-augmented view, respectively. We also develop a unified graph contrastive learning framework including similarity learning and consistency learning to make the best of two views, with our graph pooling modules being adversarially trained with respect to the graph encoder. Next, we will go into the specific components of our proposed \method{}.

\subsection{Preliminaries and Notations}\label{sec:preliminaries}

    
\smallskip\noindent\textbf{Definition 1: Graph.} Define a graph as $\mathcal{G} = (V, E, X, A)$, where $V$ represents the node set and $E$ represents the edge set. $X\in \mathbb{R}^{|V|\times d_0}$ is the node feature matrix (i.e., the $v$-th row of $X$ is the feature vector $\mathbf{x}_v$ of $v$-th node) and $A \in \mathbb{R}^{|V|\times |V|}$ denotes the adjacent matrix of the graph. 


\smallskip\noindent\textbf{Definition 2: Unsupervised Graph Representation Learning. } Given a set of unlabeled graphs $\mathcal{S} = \left\{\mathcal{G} _{1}, \cdots, \mathcal{G}_{M}\right\}$, the primary objective is to develop a graph encoder that can generate an embedding vector $\mathbf{z}_m \in \mathbb{R}^d $ for each graph $\mathcal{G}_m$, without relying on any label information. These learned graph embeddings $\{\mathbf{z}_1,\cdots  \mathbf{z}_M\}$ will be applied for downstream tasks such as graph classification.

\subsection{GNN-based Encoder}

We mainly utilize graph neural networks (GNNs) as our graph encoder due to their superior performance. GNNs typically follow the message-passing scheme to encode the structural and attributive information into node representations~\cite{gilmer2017neural}.
In particular, the propagation of the $k$-th layer of a $K$-layer GNN is described as follows:

\begin{equation}
\mathbf{h}_{v}^{(k)}= \operatorname{COM}^{(k)}_{\theta}\left(\mathbf{h}_{v}^{(k-1)}, \operatorname{AGG}^{(k)}_{\theta} \left(\left\{\mathbf{h}_{u}^{(k-1)}\right\}_{u \in \mathcal{S}(v)}\right) \right),
\end{equation}
where $\mathbf{h}_v^{(k)}$ represents the embedding of node $v$ at layer $k$, and $\mathcal{S}(v)$ is the neighbors of $v$. $\operatorname{AGG}^{(k)}_{\theta}$ is a function that aggregates information from neighbors, $\operatorname{COM}^{(k)}_{\theta}$ is a function that updates node features by combining the neighbor features and the feature of the node itself.
Finally, the graph-level representation $g_{\theta}\left(\mathcal{G}\right)$ is learned from node-level representations through a $\operatorname{READOUT}$ function, calculated as:
\begin{equation}
g_{\theta}\left(\mathcal{G} \right)=\operatorname{READOUT}\left(\left\{\mathbf{h}_{v}^{(K)}\right\}_{v \in V}\right),
\end{equation}
where $\operatorname{READOUT}$ could be a straightforward permutation invariant approach such as averaging or a more well-designed graph-level pooling function like connected layers~\cite{gilmer2017neural}. 

\subsection{Graph Pooling Module}
\begin{figure}
    \centering
    \includegraphics[width=0.4\textwidth]{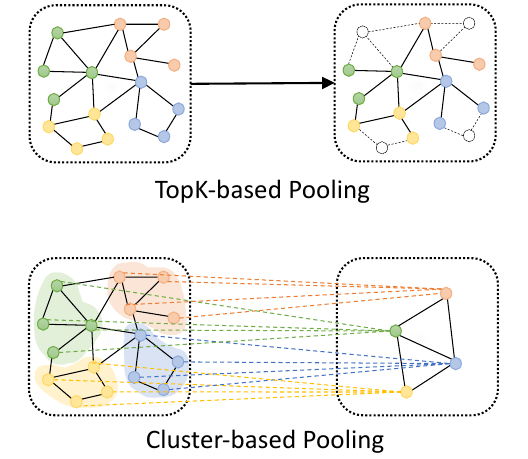}
    \caption{{Illustration of the graph pooling methods.}}
    \label{fig:pooling}
\end{figure}

Different from previous methods which introduce pre-defined augmentations to generate positive views, we leverage learnable graph pooling to generate augmented views adaptively and automatically. In formulation, we generate positive views as follows:
\begin{equation}
    \mathcal{G}_{pool} = Pool(\mathcal{G},\rho),
\end{equation}
where $\rho$ denotes the ratio of nodes to be kept. 
There are several advanced graph pooling methods to construct $Pool(\cdot, \rho)$, which can be divided into two categories as shown in Figure~\ref{fig:pooling}. Next, we introduce the details of these two categories in our framework respectively.

\smallskip\noindent\textbf{TopK-based Pooling.} In TopK-based pooling methods~\cite{lee2019self,gao2019graph}, attention mechanisms are typically adopted for adaptively selecting the nodes to be kept. In our implementation, we involve a graph encoder to generate self-attention scores $Z\in \mathbb{R}^{|V|\times 1}$ for all nodes. Then, we select the top $\lceil \rho |V|\rceil$ nodes based on the value of $Z$ to generate an index set $idx$. The calculation considers both topological information and node attributes. Finally, the pooled graph $G_{pool}$ are denoted as follows:
\begin{equation}
X_{pool}=X_{idx,:} \odot Z_{idx}, A_{pool}=A_{idx, idx},
\end{equation}
where $X_{idx,:}$ denotes the node-wise indexed feature matrix, $\odot$ denotes the broadcasted element-wise product and $A_{idx, idx}$ denotes the row-wise and column-wise indexed adjacency matrix. The pooled vertex and edge set can be inferred from $X$ and $A$. 

\smallskip\noindent\textbf{Cluster-based Pooling.} Cluster-based methods~\cite{ying2018hierarchical,ranjan2020asap} leverage graph clustering to coarsen the input graph. In our framework, we reutilize this idea and generate a cluster assignment matrix $S \in R^{|V|\times \lceil \rho |V|\rceil}$, where each row corresponds to one node while each column corresponds to one cluster. Formally, the pooled graph $\mathcal{G}_{pool}$ can be denoted as follows:
\begin{equation}
X_{pool}=S^TX,  A_{pool}=S^T AS,
\end{equation}
where $X_{i,:}$ denotes embedding of the $i$-th cluster and $A_{ij}$ denotes the the connectivity strength between cluster $i$ and cluster $j$. We generate the cluster assignment in an adaptive manner. Following \cite{ying2018hierarchical}, we generate $S$ by another learnable graph neural network with a softmax activation function.

\subsection{Contrastive Learning Framework}

In the contrastive learning framework, a critical issue is how to generate positive views for input graphs. On the one hand, we need to generate augmented views with the most removal of redundant information. Hence, their representations should be far from these input graphs for generating challenging views and providing sufficient supervision signals for contrastive tasks, which could prevent representation collapse during optimization. On the other hand, augmented graphs should preserve crucial semantic information. Since there is a trade-off between challenging positives and semantic preserving, we generate a strongly-augmented view and a weakly-augmented view for different emphases. Formally, we introduce two different ratios $\rho_1 > \rho_2$ for two augmented views $\mathcal{G}_{pool}^w = Pool(\mathcal{G}, \rho_1)$ and $\mathcal{G}_{pool}^s =Pool(\mathcal{G}, \rho_2)$. Then, we leverage different patterns to explore information from two views. 

\smallskip\noindent\textbf{Motivation for introducing Strongly-augmented View and Weakly-augmented View.} 
Motivated by ~\cite{wang2022contrastive}, we introduce two different ratios $\rho$ for two augmented views via graph pooling. We encourage to capture different semantic information from the two complementary views, and expect the patterns embedded in strong-augmented views could contribute to contrastive learning by enhancing the generalizability of learned representations. To the best of our knowledge, this could be the first work to introduce weak and strong augmentations into the graph domains.

\smallskip\noindent\textbf{Similarity Learning for Weakly-augmented Views.} Our weakly-augmented views focus on preserving semantic information, and thus we propose a contrastive task in a hard manner.
Previous approaches tend to bring different views of the same instance closer while pushing views of different samples further away~\cite{he2020momentum,chen2020simple}. In comparison, the latest contrastive learning method BYOL~\cite{grill2020bootstrap} relies only on positive views and achieves superior performance. Inspired by this, we introduce an online encoder $p_\theta$ and a target encoder $p_\phi$ sharing the same architecture. Moreover, an additional predictor $q_\theta$ is applied to the online network, which implies an asymmetric architecture. Then, we feed the original graph $\mathcal{G}$ and weakly augmented graph $\mathcal{G}_{pool}^w$ into the target encoder and online encoder respectively, producing the representations $z= p_\phi(\mathcal{\mathcal{G}})$ and $h^w= q_\theta(p_\theta(\mathcal{G}_{pool}^w))$. We minimize the cosine distance of two representations and the total loss in a batch $\mathcal{B}$ ($|\mathcal{B}|$ = $B$) is:
\begin{equation}\label{eq:sl}
    \mathcal{L}^{SL} =\frac{1}{B}\sum_{\mathcal{G} \in \mathcal{B} } 1 - \frac{z \cdot h^w}{||z||_2||h^w||_2}.
\end{equation}

\smallskip\noindent\textbf{Consistency Learning for Strongly-augmented Views.} Our strongly-augmented views are aggressive since strong augmentation could distort topological patterns and attributes. Hence, directly doing contrastive learning, which employs a ``hard" manner to achieve alignment from two views may lead to sub-optimal results. Nevertheless, strongly-augmented views can still provide some useful clues such as important motifs or subgraphs. To make the best of these clues, we develop a novel consistency learning (i.e., distributional divergence minimization) to achieve semantics consistency in a ``soft" way by considering the relation of each point to samples in the same batch. Formally, after obtaining representations of strongly-augmented graphs, i.e. $h^s=q_\theta(p_\theta(\mathcal{G}_{pool}^s))$, the similarity distribution of strongly-augmented views can be calculated by comparison with other graphs in a mini-batch as:
\begin{equation}
\mu^b=\frac{\exp \left(\cos \left(h^s, z_b \right) / \tau\right)}{\sum_{\mathcal{G}_{b'}\in \mathcal{B}} \exp \left(\cos \left(h^s, z_{b'}\right) / \tau\right)},
\end{equation}
where $z_b$ denotes the $b$-th representation in the mini-batch, $\tau$ is a temperature parameter set to be $0.5$ as in \cite{you2021graph} and $\cos(\cdot,\cdot)$ denotes the cosine similarity. In a similar way, the distribution of weakly-augmented graphs can be written as follows: 
\begin{equation}
\nu^b=\frac{\exp \left(\cos \left(h^w, z_b \right) / \tau\right)}{{\sum_{\mathcal{G}_{b'}\in \mathcal{B}}} \exp \left(\cos \left(h^w, z_{b'}\right) / \tau\right)}.
\end{equation}

Instead of hard similarity learning, we encourage the consistency between two distributions $\mu = [\mu^1, \cdots, \mu^B]$and $v = [\nu^1, \cdots, \nu^B] $ using Kullback-Leibler (KL) divergence. In formulation, the consistency learning loss is written as:
\begin{equation}\label{eq:cl}
    \mathcal{L}^{CL} = \frac{1}{B} \sum_{\mathcal{G} \in \mathcal{B} } \frac{1}{2}(D_{KL}(\mu||\nu)+D_{KL}(\nu||\mu)),
\end{equation}
where $D_{KL}(\cdot||\cdot)$ denotes the KL divergence of two distributions. Instead of directly enforcing view $h^s$ close to $z$, we propose a soft contrastive task to keep the similarity structure consistent. In this way, we explore information in strongly-augmented views while alleviating the impacts of semantic loss.

\begin{algorithm}[tb]
\caption{Training procedure of \method{}}
\label{alg:1}
\textbf{Input:} Unlabeled data $\left\{\mathcal{G} _{1}, \cdots, \mathcal{G}_{M}\right\}$, encoder parameter $\theta$, momentum parameter ${\phi}$, graph pooling module $\omega^w$ and $\omega^s$. \\
\textbf{Output}: Momentum graph encoder $g_\phi$. 

\begin{algorithmic}[1] 
\STATE Initialize $\theta$, ${\phi}$, $\omega^w$ and $\omega^s$.
\WHILE{not convergence}
    \STATE Sample $B$ graphs for $\mathcal{B}$;
    \STATE Generate $\mathcal{G}^w_{pool}$ and $\mathcal{G}^s_{pool}$ for $G \in \mathcal{S}$ ;
    \STATE Calculate the similarity learning loss by Eq.~\eqref{eq:sl}
    \STATE Calculate the consistency learning loss by Eq.~\eqref{eq:cl}
    \STATE Update $\theta$, $\omega^w$ and $\omega^s$ by Eq.~\eqref{eq:final};
    \STATE Updating $\phi$ by momentum update in Eq.~\eqref{eq:moment}.
\ENDWHILE
\end{algorithmic}
\end{algorithm}

\smallskip\noindent\textbf{Adversarial Learning for Robustness.}
Adversarial training has shown great success in improving the model robustness~\cite{kong2020flag,jiang2020smart}. In this inspirit, we leverage adversarial learning to train the graph pooling module for generating effective positive views, aiming to produce augmented graphs that are distinct from the original ones while preserving their semantic information. This way maximally enhances the optimization of contrastive learning, facilitating the learning of discriminative graph representations. Specifically, the graph pooling module is trained against the graph encoder module in an adversarial manner. The adversarial objective for weakly-augmented views is formulated in a minimax form as:
\begin{equation}\label{eq:adv}
    \min _{\theta} \max _{\omega^w} \mathcal{L}^{SL}(\theta, \omega^w),
\end{equation}
where $\omega^w$ denotes the parameters in the graph pooling module for weak augmentations. From Eq. \eqref{eq:adv}, we can observe that the graph encoder and the graph pooling module are two mutually interacted. On the one hand, the graph pooling module is trained to generate complex and robust views for effective representations. On the other hand, the graph encoder is optimized to continuously enhance the discrimination ability by minimizing the distance between input and its challenging and robust positive views. 
Unfortunately, directly minimizing the objective function as in Eq. \eqref{eq:adv} is nontrivial to find a saddle point solution. Following the optimization scheme in adversarial networks~\cite{arjovsky2017towards}, we employ a pair of gradient descent and gradient ascent applied to update parameters in the graph encoder and graph pooling module, respectively. Formally, the updating process can be formulated as:

\begin{equation}
    \left\{
    \begin{aligned}\label{eq:weak}
    \omega^w & \longleftarrow \omega^w+\eta \frac{\partial \mathcal{L}^{SL}(\theta, \omega^w)}{\partial \omega^w}\\
    \theta  & \longleftarrow \theta-\eta \frac{\partial \mathcal{L}^{SL}(\theta, \omega^w)}{\partial \theta},
    \end{aligned}
    \right.
\end{equation}
where $\eta$ denotes the learning rate. 
As for strongly-augmented views, we leverage a consistency learning objective instead of a similarity learning objective to train the graph pooling module, since we seek to release the bias bought by weakly-augmented views. As a result, the optimization scheme is defined as: 

\begin{equation}
    \left\{
    \begin{aligned}\label{eq:strong}
    \omega^s & \longleftarrow \omega^s+\eta \frac{\partial \mathcal{L}^{CL}(\theta, \omega^s)}{\partial \omega^s}\\
    \theta & \longleftarrow \theta-\eta \frac{\partial \mathcal{L}^{CL}(\theta, \omega^w, \omega^s)}{\partial \theta},
    \end{aligned}
    \right.
\end{equation}
where $\omega^s$ denotes the parameters in the graph pooling module for strong augmentations. The updated rules in Eq. \eqref{eq:weak} and \eqref{eq:strong} are summarized in a mini-batch for back-propagation updating as:
\begin{equation}
    \left\{
    \begin{aligned}\label{eq:final}
    \omega^w & \longleftarrow \omega^w+\eta \frac{\partial \mathcal{L}^{SL}(\theta, \omega^w)}{\partial \omega^w}\\
    \omega^s & \longleftarrow \omega^s+\eta \frac{\partial \mathcal{L}^{CL}(\theta, \omega^s)}{\partial \omega^s}\\
    \theta & \longleftarrow \theta- \eta \frac{\partial \mathcal{L}^{SL}(\theta, \omega^w)/\partial \theta+ \partial \mathcal{L}^{CL}(\theta, \omega^w, \omega^s)}{\partial \theta}.
    \end{aligned}
    \right.
\end{equation}
Empirical convergence can be obtained in our experiments, in accordance with the findings of other adversarial models~\cite{kong2020flag,jiang2020smart}. The momentum update is adopted in the graph encoding branch as:
\begin{equation}\label{eq:moment}
    \phi \leftarrow \gamma \phi + (1-\gamma) \theta,
\end{equation}
here, we set the momentum coefficient $\gamma$ to 0.99 following \cite{he2020momentum}. The parameters $\phi$ undergo smooth evolution through momentum updates to enhance optimization stability. The training procedure of the algorithm is shown in Algorithm \ref{alg:1}.



\section{Experiment}
\label{sec::experiment}


\subsection{Experimental Setup}

\smallskip\noindent\textbf{Datasets.}
We evaluate our proposed \method{} on two tasks: graph classification and transfer learning tasks on twelve datasets from TU datasets~\cite{morris2020tudataset} and Open Graph Benchmark (OGB) datasets~\cite{hu2020open}. For TU datasets, we adopt three bioinformatics datasets (MUTAG, PROTEINS, NCI1) and three social network datasets (IMDB-B, IMDB-M, REDDIT-M-5K) for the graph classification task. For OGB datasets, we select six molecular datasets (BBBP, ToxCast, ClinTox, BACE, HIV, MUV) for molecular property prediction under transfer learning settings. 


\smallskip\noindent\textbf{Baselines.}
We conduct a comprehensive comparison of our \method{} with three distinct groups of methods: (1) Supervised methods including GraphSage~\cite{hamilton2017inductive}, GCN~\cite{kipf2017semi}, GIN~\cite{xu2019powerful} and GAT~\cite{velivckovic2018graph}; (2) Kernel methods including Shortest Path Kernel (SP)~\cite{borgwardt2005shortest}, Graphlet Kernel (GK)~\cite{shervashidze2009efficient}, Weisfeiler-Lehman Kernel (WL)~\cite{shervashidze2011weisfeiler}; (3) Unsupervised methods including Node2vec~\cite{grover2016node2vec}, Sub2Vec~\cite{adhikari2018sub2vec}, Graph2Vec~\cite{narayanan2017graph2vec}, InfoGraph~\cite{sun2020infograph}, GraphCL~\cite{you2020graph}, JOAO~\cite{you2021graph}, AD-GCL~\cite{suresh2021adversarial}, SimGRACE~\cite{xia2022simgrace}, and GraphCLA~\cite{pu2023graph}.

\smallskip\noindent\textbf{Implementation Details.}
For our approach, we use a 2-layer GIN~\cite{xu2019powerful} as our GNN-based encoder. We set the hidden dimension of GIN as 512 and the number of training epochs as $50$. The batch size is set to 128. The ratios in graph pooling modules are set to 0.4 and 0.9 for the strongly-augmented view and weakly-augmented view, respectively. These two hyper-parameters will be discussed in Section \ref{sec:parameter_analysis}.

\begin{table*}[!t]
\caption{Performance of unsupervised learning on bioinformatics and social network classification over five runs (Averaged accuracy with standard deviation).}
\label{tab:graph_classification}
\begin{center}
\resizebox{1\textwidth}{!}{ %
\begin{tabular}{clcccccc}
\toprule
\multicolumn{2}{c}{Method} & {MUTAG} & {PROTEINS} & {NCI1} & {IMDB-B} &{IMDB-M} & {REDDIT-M-5K} \\
\midrule
\multirow{4}{*}{\begin{turn}{90}Supervised\end{turn}}
& GraphSage & 85.1 $\pm$ 7.6  & 75.3 $\pm$  2.4 & 77.7 $\pm$ 1.5  & 72.3 $\pm$ 5.3 & 50.9 $\pm$ 2.2 & 43.8 $\pm$ 3.2  \\
& GCN & 85.6 $\pm$ 5.8 &  75.2 $\pm$ 3.6  & 80.2 $\pm$ 2.0 & 74.0 $\pm$ 3.4 & 51.9 $\pm$ 3.8 & 20.0 $\pm$ 0.0       \\
& GIN & \textbf{89.4 $\pm$ 5.6} &  \textbf{76.2 $\pm$ 2.8} & \textbf{82.7 $\pm$ 1.7} & \textbf{75.1 $\pm$ 5.1} & \textbf{52.3 $\pm$ 2.8} & \textbf{57.6 $\pm$ 1.5}   \\
& GAT & \textbf{89.4 $\pm$ 6.1} &  74.7 $\pm$ 4.0 & 66.6 $\pm$ 2.2 & 70.5 $\pm$ 2.3 & 47.8 $\pm$ 3.1 & 45.9 $\pm$ 0.1   \\
\midrule
\multirow{3}{*}{\begin{turn}{90}Kernel\end{turn}}  
& SP  & 85.2 $\pm$ 2.4 &  $-$   & 73.5 $\pm$ 0.1  & 55.6 $\pm$ 0.2 & 38.0 $\pm$ 0.3 & 39.6 $\pm$ 0.2   \\
& GK & 81.7 $\pm$ 2.1 &  $-$   & 66.0 $\pm$ 0.1  & 65.9 $\pm$ 1.0 & 43.9 $\pm$ 0.4 & 41.0 $\pm$ 0.2   \\
& WL  & 80.7 $\pm$ 3.0 &  72.9 $\pm$ 0.6 & $-$  & 72.3 $\pm$ 3.4 & 47.0 $\pm$ 0.5 & 46.1 $\pm$ 0.2  \\
\midrule
\multirow{9}{*}{\begin{turn}{90}Unsupervised\end{turn}}
& Node2Vec  & 72.6 $\pm$ 10.2 &  57.5 $\pm$ 3.6 & 54.9 $\pm$ 1.6 & 50.2 $\pm$ 0.9  & 36.0 $\pm$ 0.7  & $-$  \\
& Sub2Vec   & 61.1 $\pm$ 15.8 & 53.0 $\pm$ 5.6 & 52.8 $\pm$ 1.5 & 55.3 $\pm$ 1.5 & 36.7 $\pm$ 0.8  & 36.7 $\pm$ 0.4   \\
& Graph2Vec & 83.2 $\pm$ 9.6  &  73.3 $\pm$ 2.1 & 73.2 $\pm$ 1.8 &  71.1 $\pm$ 0.5 & 46.3 $\pm$ 1.4 & 47.9 $\pm$0 .3   \\
& InfoGraph & 89.0 $\pm$ 1.1  &  74.4 $\pm$ 0.3  & 76.2 $\pm$ 1.1 & 71.1 $\pm$ 0.9 & 49.7 $\pm$ 0.5 & 53.5 $\pm$ 1.0  \\ 
& GraphCL   & 86.8 $\pm$ 1.3  & 74.4 $\pm$ 0.5 & 77.9 $\pm$ 0.4 & 71.1 $\pm$ 0.4 & 48.5 $\pm$ 0.6  & 56.0 $\pm$ 0.3  \\
& JOAO & 87.3 $\pm$ 1.0  &  74.6 $\pm$ 0.4 & 78.1 $\pm$ 0.5 &  70.2 $\pm$ 3.1 & $-$  & 55.7 $\pm$ 0.6  \\
& AD-GCL & 89.3 $\pm$ 1.5 & 73.6 $\pm$ 0.7 & 69.7 $\pm$ 0.5 &  71.6 $\pm$ 1.0 & 49.0 $\pm$ 0.5 & 54.9 $\pm$ 0.4  \\
& SimGRACE &89.1 $\pm$ 1.4 &74.9 $\pm$ 0.7 &79.1 $\pm$ 0.5 &71.6 $\pm$ 0.7 &48.7 $\pm$ 0.7 &55.9 $\pm$ 0.4  \\
& GraphCLA &89.3 $\pm$ 0.4 &74.5 $\pm$ 0.6 &73.0 $\pm$ 0.6 &72.3 $\pm$ 0.5 &49.5 $\pm$ 0.4 &$-$  \\
\cmidrule{2-8}
& \method{}-TopK (Ours) & \textbf{89.9 $\pm$ 0.7}  & \textbf{75.1 $\pm$ 0.4} & 79.1 $\pm$ 0.6 & 73.5 $\pm$ 0.7  &  51.4 $\pm$ 0.6  & \textbf{56.3 $\pm$ 0.2}\\
& \method{}-Cluster (Ours) &  89.5 $\pm$ 1.2  & 74.7 $\pm$ 0.5 &\textbf{79.5 $\pm$ 0.4}  & \textbf{73.8 $\pm$ 1.1} & \textbf{51.7 $\pm$ 0.5}  & 55.9 $\pm$ 0.4  \\
\bottomrule
\end{tabular}
}
\end{center}
\end{table*}

\subsection{Experimental Results}

As shown in Table \ref{tab:graph_classification}, we evaluate the effectiveness of our \method{} for graph classification, compared to various baselines. We can draw the following conclusions:


\begin{itemize}
\item Overall, from the results, it can be observed that our proposed model \method{} shows superior performance across all six datasets. \method{} consistently performs better than other unsupervised baselines by a significant margin. The strong performance demonstrates the effectiveness of the proposed multi-scale pooling framework for effective graph contrastive learning. 
\item A general observation is that supervised algorithms still have the highest performance. Interestingly, even compared with the supervised ones, our approach \method{} achieves competitive performance in 5 out of 6 datasets and outperforms supervised results on dataset MUTAG. Moreover, among all the supervised algorithms, we can see that GIN consistently outperforms other GNN models on all datasets, which verifies the superiority of GIN with strong representation capability. This justifies the reason why we choose GIN as the base GNN-based encoder.
\item The performance of traditional kernel methods is inferior to most unsupervised methods, which suggests that these methods may be ineffective in capturing effective information of the graph topology and node attributes. Moreover, the features derived from kernel methods are typically heuristic, which leads to worse generalization ability and sub-optimal performance.
\item By integrating the idea of contrastive learning into GNNs, recent state-of-the-art methods (InfoGraph, GraphCL, JOAO, AD-GCL, SimGRACE, and GraphCLA) have obtained high enough performance, which pushes away the other unsupervised baselines (Node2Vec, Sub2Vec, Graph2Vec), sufficiently showing the superiority of instance discrimination principle in the contrastive learning.
\item Among two variants based on different graph pooling techniques, we can see that \method{}-TopK and \method{}-Cluster stand out as two robust variants. They achieve top-tier or competitive performance across all datasets. Compared to existing state-of-the-arts, their superior results validate the effectiveness of our framework, which explores learnable graph pooling to derive augmented views in an adversarial manner.
\end{itemize}



\begin{figure}[t]
\centering
\resizebox{0.95\textwidth}{!}{
    \subfloat[MUTAG]{
        \includegraphics[width=0.28\textwidth]{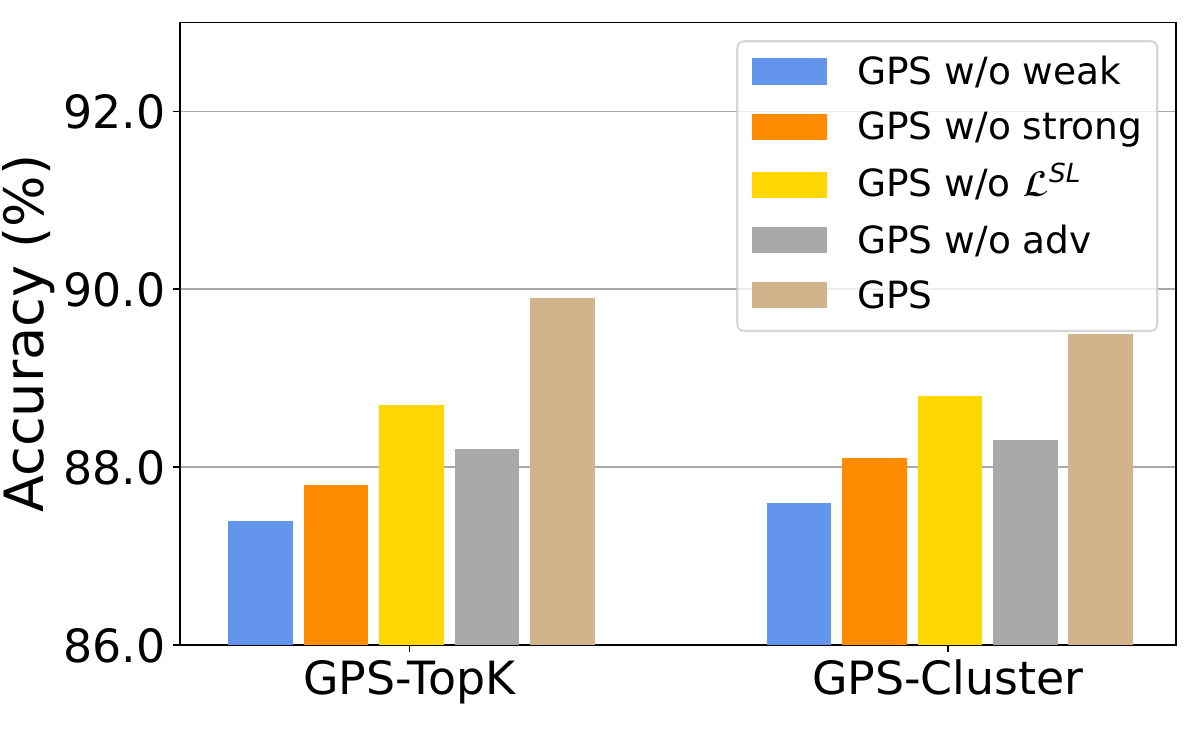}}
    \hfill
    \subfloat[PROTEINS]{
        \includegraphics[width=0.28\textwidth]{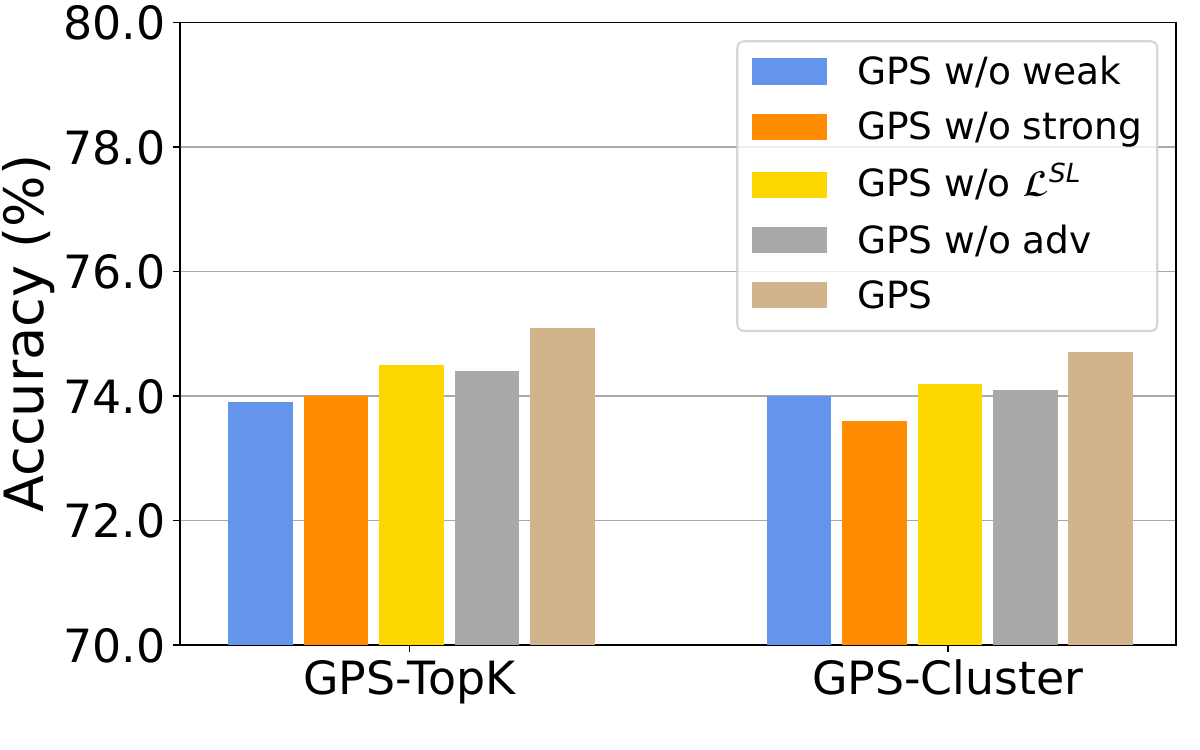}}
    \hfill
    \subfloat[NCI1]{
        \includegraphics[width=0.28\textwidth]{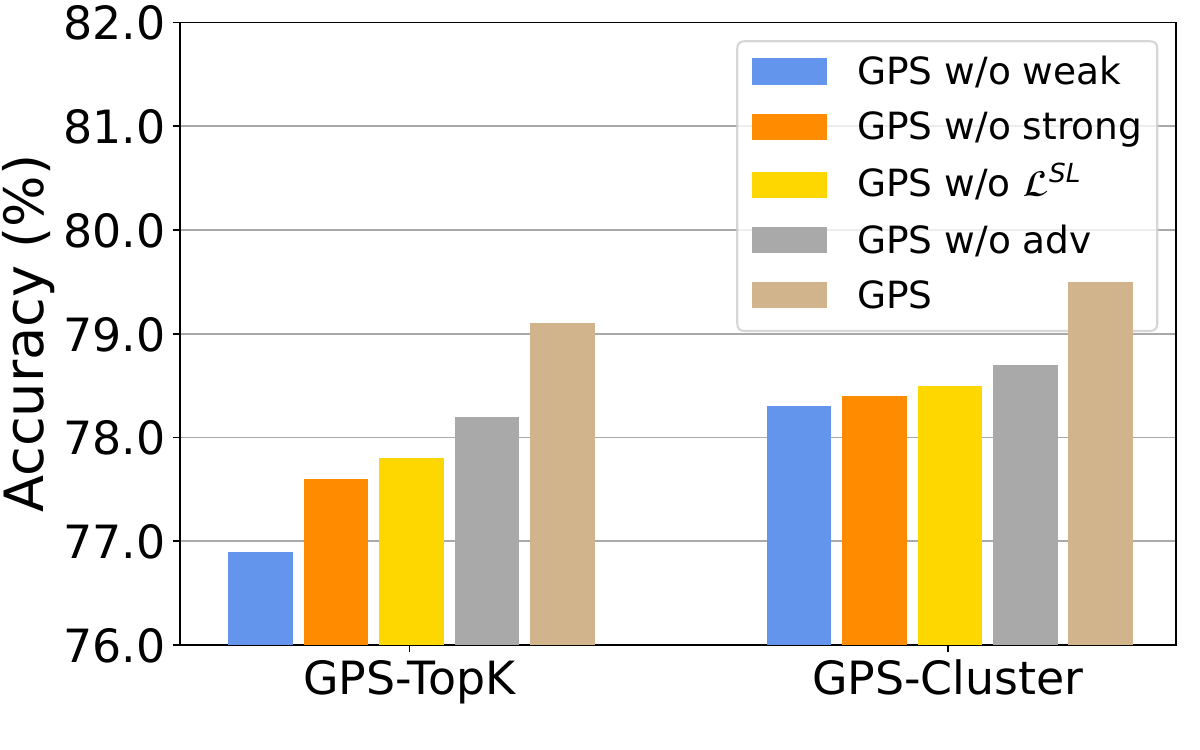}}
}
\resizebox{0.95\textwidth}{!}{
    \subfloat[IMDB-B]{
        \includegraphics[width=0.28\textwidth]{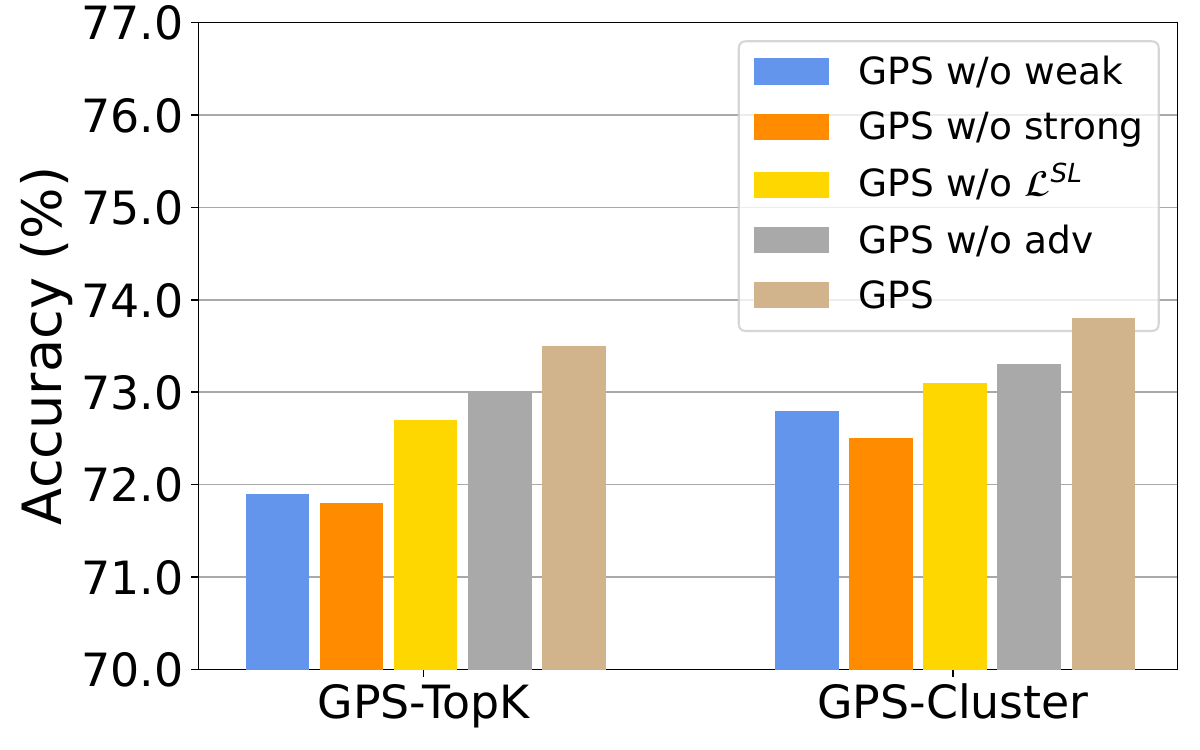}}
    \hfill
    \subfloat[IMDB-M]{
        \includegraphics[width=0.28\textwidth]{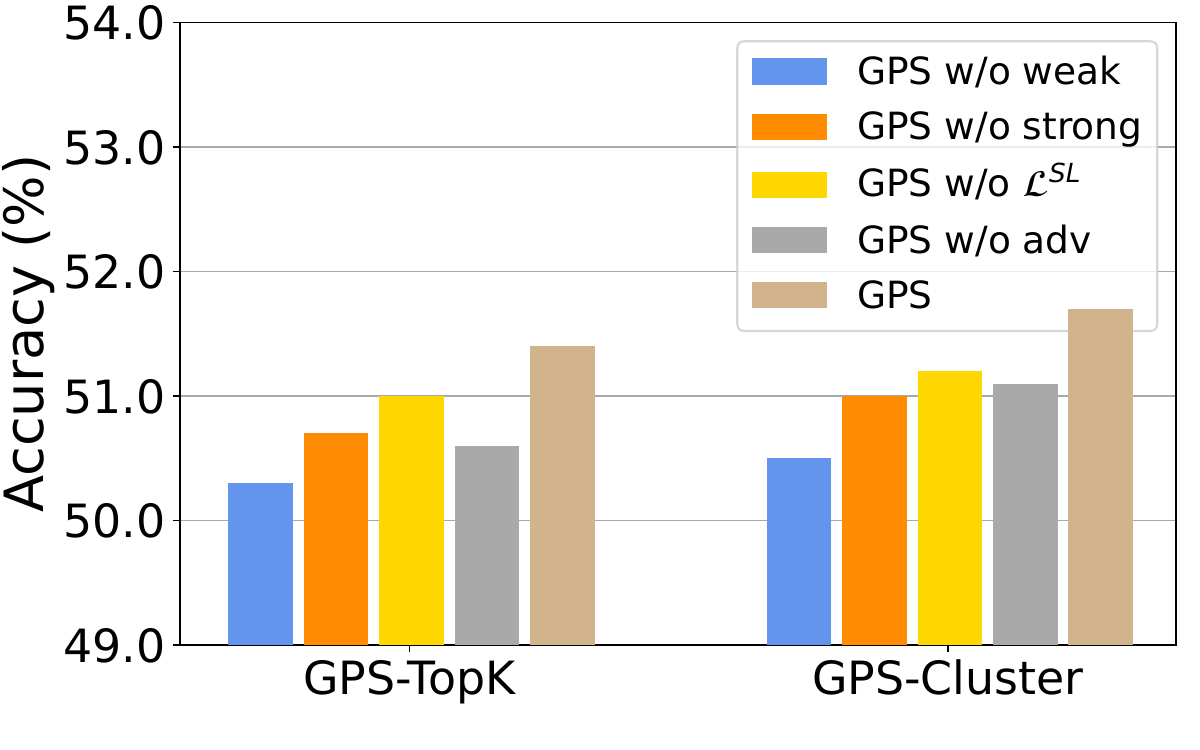}}
    \hfill
    \subfloat[REDDIT-M-5K]{
        \includegraphics[width=0.28\textwidth]{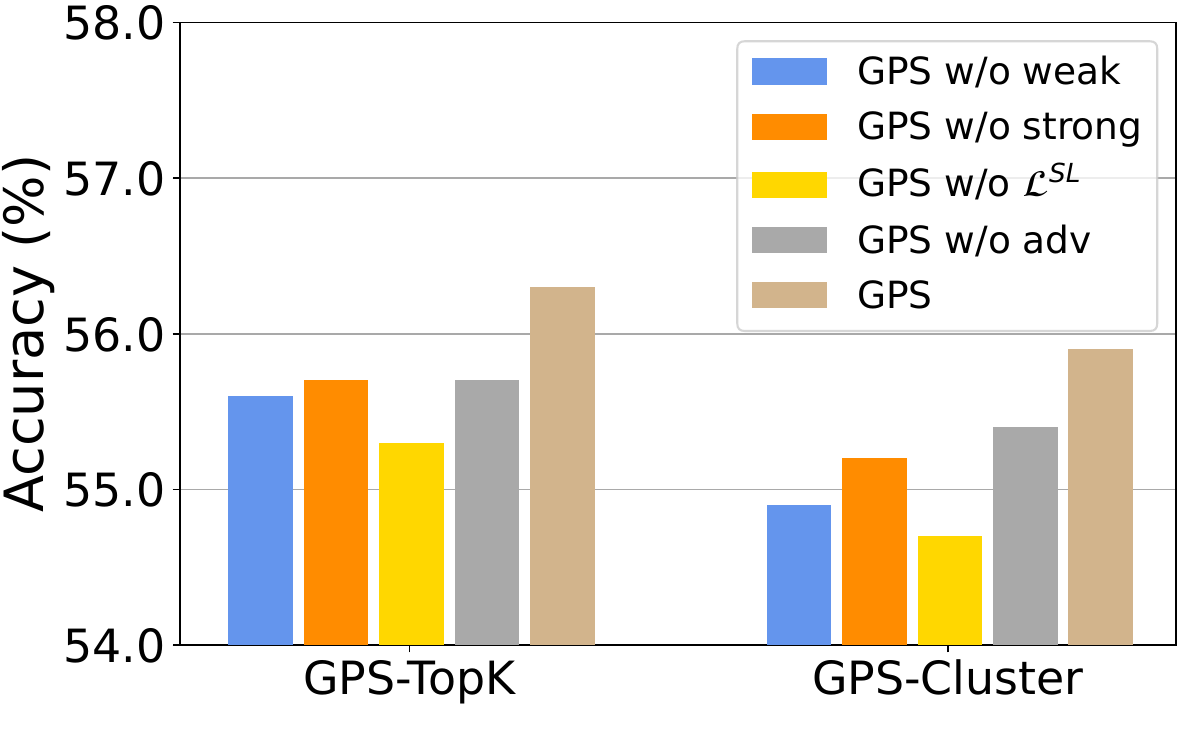}}
}    
\caption{Performance of ablation study of several model variants (in $\%$) on all six datasets.}
\label{fig:ablation}
\end{figure}

\subsection{Ablation Study}

Then, we compare \method{} with its three variants to validate the effectiveness of each component.
\begin{itemize}
\item \textbf{\method{} w/o weak}: We remove the weakly-augmented view and train the model with similarity learning using the strongly-augmented view since consistency learning requires both views. 
\item \textbf{\method{} w/o strong} (\textbf{$\mathcal{L}^{CL}$}): We remove the strongly-augmented view and the model is simply trained with similarity learning using the weakly-augmented view. 
\item \textbf{\method{} w/o $\mathcal{L}^{SL}$}: We remove the similarity learning loss and the model is simply trained with consistency learning using both views.
\item \textbf{\method{} w/o adv}: We remove the adversarial learning in the graph pooling modules. The pooling modules are updated with gradient descent along with the encoder. 
\end{itemize}



We compare the performance of different variants and then plot the results in Figure \ref{fig:ablation}. From the figure, we can draw the following conclusions. First, the results of \method{} are consistently better than all the other four variants, indicating that both our multi-scale graph pooling and adversarial learning are effective for graph contrastive learning. Second, the results of \method{} w/o weak and \method{} w/o strong are usually inferior to \method{} w/o adv on most datasets, which verifies the usefulness of the two view augmentations. Third, \method{} w/o strong is generally better than \method{} w/o weak on most datasets, which implies that weakly-augmented views as well as similarity learning play a more important role in this framework. Fourth, we observe that removing the strongly-augmented view is equivalent to removing $\mathcal{L}^{CL}$. It can be noticed that regardless of which loss is removed (\method{} w/o strong ($\mathcal{L}^{CL}$) or \method{} w/o $\mathcal{L}^{SL}$), the performance of our proposed method deteriorates significantly, demonstrating the significance of our proposed diverse losses. Additionally, \method{} w/o $\mathcal{L}^{SL}$ outperforms $\mathcal{L}^{CL}$ in five out of six datasets, highlighting the importance of emphasizing strongly-augmented and weakly-augmented views for learning discriminative graph representations.

\begin{figure}[t]
\centering
\includegraphics[width=0.65\textwidth]{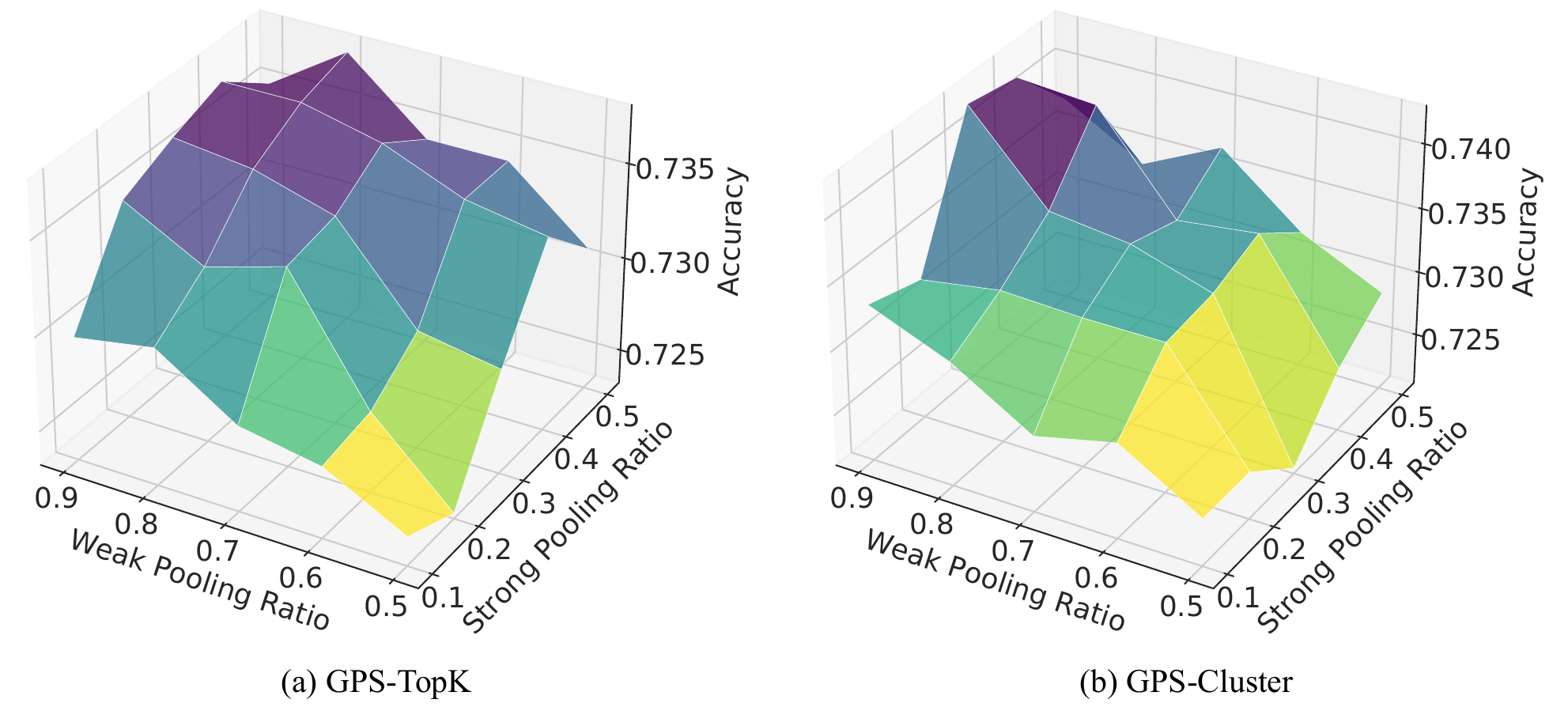}
\caption{Analysis of graph pooling ratio on IMDB-B.}
\label{fig:sensitivity-analysis}
\end{figure}

\subsection{Sensitivity Analysis}
\label{sec:parameter_analysis}

In this section, we investigate the sensitivity of parameters graph pooling ratio $\rho$ and batch size $B$.

\textbf{Analysis of graph pooling ratio.}
We test the effect of the graph pooling ratio $\rho$, which controls the ratio of the augmented graph. We vary $\rho_1$ and $\rho_2$ as $\{0.1, 0.2, 0.3, 0.4, 0.5\}$ and $\{0.5, 0.6, 0.7, 0.8, 0.9\}$, respectively. The results of our two variants on IMDB-B are shown in Figure~\ref{fig:sensitivity-analysis}. We can observe that for two variants, generally, with the decrease of $\rho_1$ or $\rho_2$ while the other ratio is fixed, the performance tends to decrease slowly. Maybe the reason is that the small ratio of graph pooling prone to distort topological patterns and attributes. However, note that for \method{}-TopK, the performance difference caused by different parameter combinations is less than 0.01, and for \method{}-Cluster, the performance is relatively stable when the parameters are not too large or small, as shown in the plateau in the Figure~\ref{fig:sensitivity-analysis} (b). We conjecture that it is beneficial to performance via generating augmented views with the removal of redundant information and preserving semantics in an adversarial way. We hence conclude that our proposed framework \method{} is generally insensitive to these parameters, demonstrating the robustness to hyperparameter tuning and easing the parameter selection for our framework.

\begin{figure}
\centering
\includegraphics[width=0.65\textwidth]{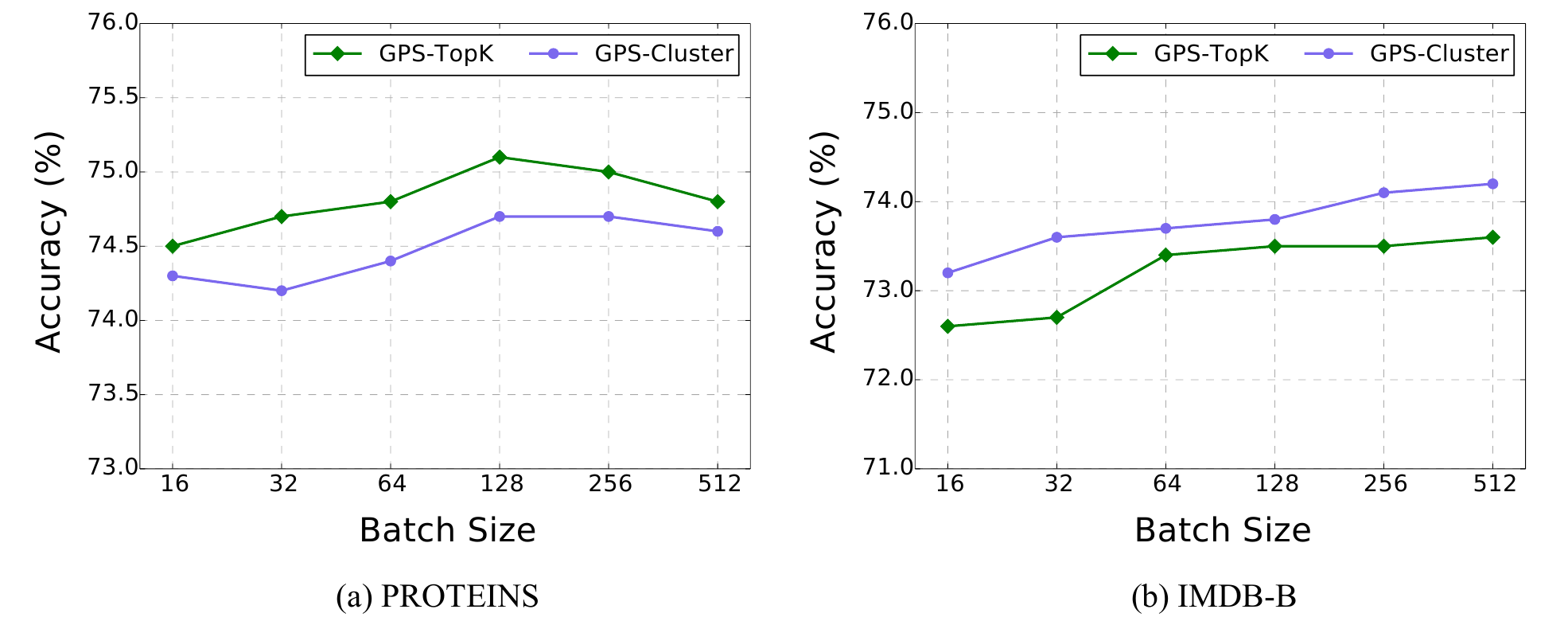}
\caption{Analysis of batch size on PROTEINS and IMDB-B.}
\label{fig:batch_size}
\end{figure}

\textbf{Analysis of batch size.}
Next, we evaluate the effect of the batch size $B$, and vary it in the range of $\{16, 32, 64, 128, 256, 512\}$. The results are shown in Figure~\ref{fig:batch_size}. It can be seen that (i) for PROTEINS, with the increase of $B$, the performance tends to first increase and then decrease. A too-small $B$ would lead to a lack of intra-batch sample diversity and fail to provide an effective similarity distribution while a large $B$ may introduce too many noise samples. 
(ii) For IMDB-B, we can observe that an increasing batch size consistently enhances performance. This is because a sufficiently large batch can more effectively represent the entire dataset, encompassing a wider range of diverse samples to facilitate the learning of discriminative representations for the target samples. It's worth noting that an excessively large batch size could potentially lead to issues related to space complexity.

\begin{table*}[!t]
\caption{The clustering performance on four graph property prediction benchmarks.}
\label{tab:clustering}
\begin{center}
\resizebox{1\textwidth}{!}{ %
\begin{tabular}{lcccccccccccc}
\toprule
Method & \multicolumn{3}{c}{DD} & \multicolumn{3}{c}{IMDB-B} & \multicolumn{3}{c}{REDDIT-B} & \multicolumn{3}{c}{REDDIT-M-12K}  \\
\midrule
Metrics &NMI &ACC &ARI &NMI &ACC &ARI &NMI &ACC &ARI &NMI &ACC &ARI \\
\midrule
InfoGraph &0.008 &0.558 &-0.006  &0.041 &0.538 &0.005 &0.016 &0.508 &0.000 &0.045 &0.205 &0.003 \\
GraphCL &0.019 &0.573 &-0.009  &0.046 & 0.545 &0.008 &0.033 &0.519 &0.001 &0.096 &0.181 &0.021 \\
CuCo &0.012 &0.562 &-0.010 &0.001 &0.507 &0.000 &0.018 &0.510 &0.000 &0.003 &0.192 &0.002 \\
JOAO &0.012 &0.578 &-0.004 &0.042 &0.543 &0.008 &0.034 &0.520 &0.001 &0.003 &0.183 &0.001 \\
RGCL &0.014 &0.565 &-0.009 &0.047 &0.546 &0.007 &0.017 &0.509 &0.001 &0.003 &0.092 &0.001 \\
SimGRACE &0.001 &0.589 &0.003 &\textbf{0.049} &0.559 &0.007 &0.024 &0.513 &0.001 &0.062 &0.210 &0.005 \\
\midrule
\method{} &\textbf{0.020} &\textbf{0.594} &\textbf{0.004} &0.048 &\textbf{0.0565} &\textbf{0.009} &\textbf{0.035} &\textbf{0.523} &\textbf{0.002} &\textbf{0.113} &\textbf{0.220} &\textbf{0.035} \\
\bottomrule
\end{tabular}
}
\end{center}
\end{table*}

\subsection{Graph-Level Clustering}

To further demonstrate the discrimination of the learned graph representations, we conduct the experiment of graph-level clustering~\cite{ju2023glcc} on four datasets, including one DD, IMDB-B, REDDIT-B, and REDDIT-M-12K. We compare our \method{} with several competitive baselines: InfoGraph~\cite{sun2020infograph}, GraphCL~\cite{you2020graph}, CuCo~\cite{chu2021cuco}, JOAO~\cite{you2021graph}, RGCL~\cite{li2022let} and SimGRACE~\cite{xia2022simgrace}. Here we
adopt three widely-used evaluation indicators to measure the clustering performance: Normalized Mutual Information (NMI)~\cite{strehl2002cluster}, clustering Accuracy (ACC)~\cite{li2006relationships} and Adjusted Rand Index (ARI)~\cite{hubert1985comparing}. These evaluation indicators cover various aspects of clustering outcomes. NMI and ACC have a range of $[0,1]$, whereas ARI ranges in $[-1,1]$. Higher values indicate better performance across all three evaluation indicators.

The quantitative results of graph-level clustering are reported in Table~\ref{tab:clustering}, it can be observed that our proposed \method{} consistently demonstrates superior performance compared to other graph contrastive learning approaches across all four datasets under three evaluation indicators. This showcases the exceptional effectiveness of our framework in graph-level clustering. This might be attributed to our multi-scale augmented views, which capture complementary information and learn more discriminative representations through adversarial learning, thereby better serving the clustering task.

\begin{table*}[t]
\caption{Performance of transfer learning on molecular property prediction over five runs (ROC-AUC with standard deviation).}
\label{tab:transfer_learning}
\centering
\resizebox{1\textwidth}{!}{ %
\begin{tabular}{lcccccc|cc}
\toprule
Pre-Train Dataset   & \multicolumn{6}{c}{\begin{tabular}[c]{@{}c@{}} ZINC15 2M\end{tabular}} \\ \midrule
Fine-Tune Dataset   & BBBP  & ToxCast  & ClinTox  & BACE & HIV & MUV & Avg. & Ranks  \\ 
\midrule
No Pre-Train & 65.8 $\pm$ 4.5  & 63.4 $\pm$ 0.6 & 58.0 $\pm$ 4.4 & 70.1 $\pm$ 5.4 & 75.3 $\pm$ 1.9 & 71.8 $\pm$ 2.5  & 67.4  & 10  \\
EdgePred~\cite{hu2019strategies} & 67.3 $\pm$ 2.4  & 64.1 $\pm$ 0.6  & 64.1 $\pm$ 3.7 & 79.9 $\pm$ 0.9 & 76.3 $\pm$ 1.0 & 74.1 $\pm$ 2.1  & 71.0  & 9  \\
AttrMasking~\cite{hu2019strategies}  & 64.3 $\pm$ 2.8   & 64.2 $\pm$ 0.5  & 71.8 $\pm$ 4.1 & 79.3 $\pm$ 1.6 & 77.2 $\pm$ 1.1 & 74.7 $\pm$ 1.4  & 71.9  & 5 \\
ContextPred~\cite{hu2019strategies}  & 68.0 $\pm$ 2.0  & 63.9 $\pm$ 0.6  & 65.9 $\pm$ 3.8 & 79.6 $\pm$ 1.2 & 77.3 $\pm$ 1.0  & \textbf{75.8 $\pm$ 1.7} &  71.8 & 6  \\
\midrule
GraphPartition~\cite{you2020does}  & 70.3 $\pm$ 0.7  & 63.2 $\pm$ 0.3  & 64.2 $\pm$ 0.5 & 79.6 $\pm$ 1.8  & 77.1 $\pm$ 0.7 & 75.4 $\pm$ 1.7  & 71.6  & 7\\
InfoGraph~\cite{sun2020infograph}  & 68.8 $\pm$ 0.8  & 62.7 $\pm$ 0.4  & 69.9 $\pm$ 3.0 & 75.9 $\pm$ 1.6  & 76.0 $\pm$ 0.7  & 75.3 $\pm$ 2.5  &  71.4 & 8  \\
GraphCL~\cite{you2020graph}  & 69.7 $\pm$ 0.7  & 62.4 $\pm$ 0.6  & 76.0 $\pm$ 2.7 & 75.4 $\pm$ 1.4  & 78.5 $\pm$ 1.2  & 69.8 $\pm$ 2.7  &  72.0 & 4  \\ 
JOAO~\cite{you2021graph} & 70.2 $\pm$ 1.0  & 62.9 $\pm$ 0.5  & 81.3 $\pm$ 2.5 & 77.3 $\pm$ 0.5 & 76.7 $\pm$ 1.2 & 71.7 $\pm$ 1.4  &  73.4 & 3\\
AD-GCL~\cite{suresh2021adversarial} & 70.0 $\pm$ 1.0  & 63.1 $\pm$ 0.7  & 79.8 $\pm$ 3.5 & 78.5 $\pm$ 0.8 & 78.3 $\pm$ 1.0 & 72.3 $\pm$ 1.6  &  73.7 & 2\\
\midrule
\method{}-TopK  & \textbf{71.5 $\pm$ 0.9}  & \textbf{64.4 $\pm$ 0.3} & \textbf{82.1 $\pm$ 2.9} & \textbf{80.1 $\pm$ 0.8} & \textbf{79.0 $\pm$ 1.1} & \multicolumn{1}{c|}{75.6 $\pm$ 1.7}  & \textbf{75.5} & \textbf{1} \\
\bottomrule
\end{tabular}%
}
\end{table*}

\subsection{Transfer Learning}

In this section, we evaluate the generalization of our proposed method on molecular property prediction for transfer learning. Following \cite{hu2019strategies}, our model is pre-trained on a large-scale ZINC15 dataset (two million unlabeled molecules) and later fine-tuned on six Open Graph Benchmark (OGB)~\cite{hu2020open} datasets to test out-of-distribution performance.
Here, we only consider \method{}-TopK for illustration. We adopt four common pre-training strategies (No Pre-Train, EdgePred, AttrMasking, and ContextPred in~\cite{hu2019strategies}) and five state-of-the-art techniques (GraphPartition~\cite{you2020does}, InfoGraph~\cite{sun2020infograph}, GraphCL~\cite{you2020graph}, JOAO~\cite{you2021graph}, and AD-GCL~\cite{suresh2021adversarial}) to study the transferability of the various pre-training strategies. 

The results are shown in Table~\ref{tab:transfer_learning}. It can be observed that \method{}-TopK significantly outperforms various baselines in five out of the six datasets, and achieve Top-1 performance in terms of average ROC-AUC among ten baselines, which fully shows the excellent generalization capacity of our framework. Moreover, it is worth mentioning that compared to the No Pre-Train, our method improves 41.6$\%$ and 14.3$\%$ on ClinTox and BACE respectively, indicating the effectiveness of the discrimination ability of the contrastive learning principle. Compared with the stronger baselines GraphCL~\cite{you2020graph}, JOAO~\cite{you2021graph} and AD-GCL~\cite{suresh2021adversarial}, we can see those different methods may have their own preference for different datasets due to their specific characteristics such as binding affinity, toxicity and adverse reactions. However, our method can consistently outperform these baselines in most datasets, indicating the effectiveness of the multi-scale pooling. The above results show that our method \method{} can learn effective graph-level representations which achieve superior out-of-distribution performance.

\subsection{Semi-Supervised Learning}

Lastly, we evaluate our proposed model for semi-supervised learning on two large-scale OGB datasets~\cite{morris2020tudataset} ogbg-ppa and ogbg-code to test the scalability of our framework. The ogbg-ppa dataset, which consists of 158,100 proteins, is extracted from the protein-protein association networks of 1,581 different species. The ogbg-code dataset is a collection of Abstract Syntax Trees obtained from 452,741 Python method definitions. Here, we only consider \method{}-TopK for illustration. Our model is pre-trained on one dataset using self-supervised learning and later fine-tuned based on 3$\%$ and 10$\%$ label supervision on the same dataset following the setting in \cite{you2021graph} and compare it with GraphCL~\cite{you2020graph} and JOAO~\cite{you2021graph}.  

The results are reported in Table~\ref{tab:results_ogb}. From the table, we can see that our \method{}-TopK significantly outperforms all the baselines on two large-scale OGB datasets, which again demonstrates the strength and scalability of our proposed model. Maybe the reason is that GraphCL and JOAO adopt the empirically pre-defined rules for augmentation selection while our framework leverage learnable graph pooling to automatically provide effective augmented views.

\begin{table}[t]
\caption{Performance of semi-supervised learning on large-scale OGB datasets. (Accuracy on ogbg-ppa, F1 on ogbg-code at $3\%$ and $10\%$ label rate respectively.)}
\label{tab:results_ogb}
\centering
\resizebox{0.45\textwidth}{!}{ %
\begin{tabular}{@{}llcc@{}}
\toprule
{Rate}   & {Methods} & {ogbg-ppa} & {ogbg-code}  \\
\midrule 
\multirow{3}*{3\%} & GraphCL  & $ 44.3 \pm 5.2 $ & $ 12.0 \pm 0.3 $ \\
& JOAO  & $ 47.8\pm 4.6 $ & $ 11.7 \pm 0.6 $  \\
\cmidrule{2-4}
& \method{}-TopK &  \textbf{49.6 $\pm$ 3.9}  &   \textbf{ 13.2 $\pm$ 0.4 }  \\
\midrule
\multirow{3}*{10\%} & GraphCL  & $ 55.8 \pm 0.9 $ & $ 20.9 \pm 0.3 $ \\
& JOAO  & $ 60.1\pm1.2 $ & $ 21.4 \pm 0.5 $  \\
\cmidrule{2-4}
& \method{}-TopK &  \textbf{63.2 $\pm$ 1.1}  &   \textbf{23.1 $\pm$ 0.5}  \\
\bottomrule
\end{tabular}
}
\end{table}




\section{Conclusion}
\label{sec::conclusion}

In this study, we explore self-supervised graph representation learning by presenting a novel framework Graph Pooling ContraSt (\method{}). Specifically, \method{} leverages learnable graph pooling to automatically generate multi-scale positive views, which emphasize preserving semantics and providing challenging positives via strongly-augmented view and weakly-augmented view, respectively. Moreover, we develop a joint contrastive learning framework that incorporates both views to explore similarity learning and consistency learning, where our graph pooling modules are adversarially trained with the encoder for robustness and efficiency. Extensive experiments well showcase the superiority of our proposed \method{} over state-of-the-art baselines on twelve real-world datasets. 

\Acknowledgements{This paper is partially supported by National Key Research and Development Program of China with Grant No. 2023YFC3341203, the National Natural Science Foundation of China (NSFC Grant Numbers 62306014 and 62276002) as well as the China Postdoctoral Science Foundation with Grant No. 2023M730057.}


\bibliographystyle{splncs04}
\bibliography{mybibliography}



\end{document}